\documentclass[sigconf, nonacm, anonymous=false, table]{acmart}
\AtBeginDocument{%
  }

\setcopyright{acmlicensed}
\copyrightyear{2018}
\acmYear{2018}
\acmDOI{XXXXXXX.XXXXXXX}
\acmConference[Conference acronym 'XX]{Make sure to enter the correct
  conference title from your rights confirmation email}{June 03--05,
  2018}{Woodstock, NY}
\acmISBN{978-1-4503-XXXX-X/2018/06}

\newcommand{\etal}{\textit{et al.}}
\usepackage[]{algpseudocode}      
\usepackage{algorithm}
\usepackage{bm,bbm}
\usepackage{cleveref}
\usepackage{booktabs}   
\usepackage{multirow}   
\usepackage{makecell}   
\usepackage{graphicx}   
\usepackage{xcolor}     
\theoremstyle{definition}
\newtheorem{mydefinition}{\textbf{Definition}}
\newtheorem{myproblem}{\textbf{Problem}}

\usepackage{enumitem}
\usepackage{tabularx}
\begin{document}
\pagestyle{empty}

\title{LLM-Augmented Bayesian Optimization with Adaptive Policies for Chiplet Design Space Exploration}

\author{
    Hang Liu$^{1,2}$, Zhengfei Chen$^{1}$, Zhengrui Chen$^{1,2}$, Hao Geng$^{3}$, Yu Li$^{1}$, Qi Sun$^{1,2,*}$ \\
    $^1$Zhejiang University, Hangzhou, China;
    $^2$ChipFlux, Hangzhou, China;\\
    $^3$ShanghaiTech University, Shanghai, China;\\
    $^*$Corresponding authors, email: qisunchn@zju.edu.cn
    \vspace{-1ex}
}


\begin{abstract}
Chiplet technology is a key approach for extending Moore's Law.
However, its design process is constrained by two primary challenges: the scarcity of early-stage platforms for rapidly evaluating chiplet-based heterogeneous systems, and the complex relationships between design parameters and optimization metrics.
To address these challenges, we propose a synergistic framework that integrates a Large Language Model (LLM) with Bayesian Optimization (BO), enabling architectural-level simulation and automated optimization for the 2.5D/3D CPU-GPU heterogeneous chiplets.
We extend the open-source chiplet simulator to evaluate 3D designs by supporting graph topologies.
Further, to enhance the exploration efficiency of BO, we introduce an LLM that generates high-quality policies to guide the optimization process, enabled by our proposed adaptive directional policy reasoning (ADPR) and an exploration gain-aware fine-tuning method.
Finally, we use the policy-guided BO to perform an automated and efficient exploration of the complex chiplet design space.
Experiments on 2.5D and 3D chiplet reveal that our framework improves design space exploration efficiency.

\end{abstract}

\maketitle

\section{Introduction}

In the post-Moore era \cite{chiplet}, where traditional scaling faces diminishing returns and rising costs, chiplet-based heterogeneous systems have emerged as a key paradigm to overcome monolithic SoC limitations \cite{chiplet_survey}.
This approach allows flexible integration of modular dies from diverse process nodes and functionalities, yielding superior performance at lower manufacturing costs. However, such high modularity and heterogeneity introduce inherent challenges in design complexity.

\begin{figure}[th]
    \centering
    \includegraphics[width=0.95\linewidth]{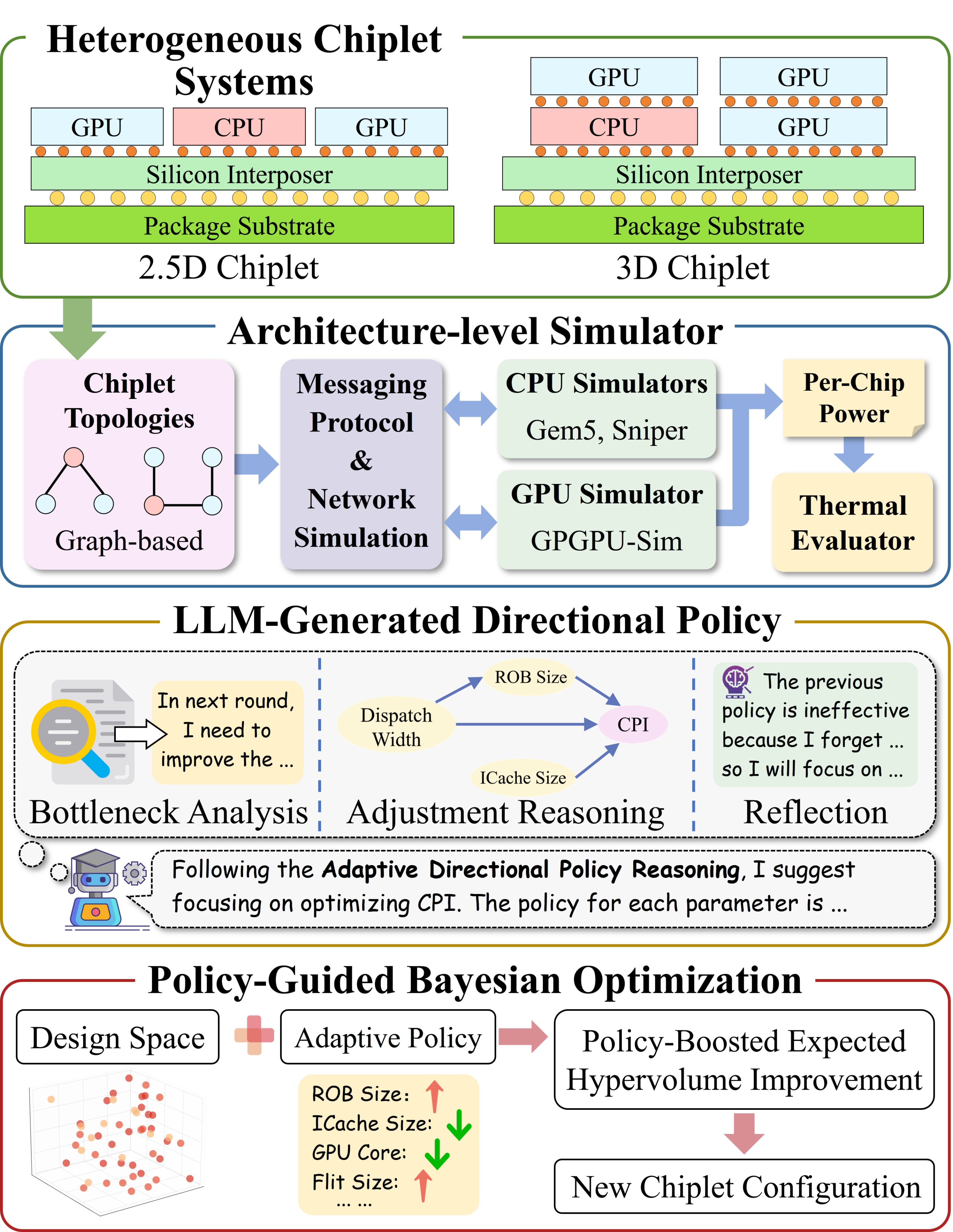}
    \Description{Overview of our automated chiplet design framework, synergizing an extended simulator for architectural evaluation of 2.5D/3D chiplets with LLM-guided Bayesian Optimization for efficient design optimization.}
    \caption{Overview of our automated chiplet design framework, synergizing an extended simulator for architectural evaluation of 2.5D/3D chiplets with LLM-guided Bayesian Optimization for efficient design optimization.}
    \label{fig:flow}
\end{figure}

A critical bottleneck in chiplet development is the lack of adequate early-stage evaluation tools for CPU-GPU heterogeneous chiplets.
Existing simulators, such as HeteroGarnet \cite{garnet} and BookSim2 \cite{booksim}, primarily focus on interconnect modeling and do not support the holistic simulation of  2.5D/3D heterogeneous chiplets.
Therefore, we need a method that provides rapid and reliable performance evaluation early in the design cycle to guide design directions and avoid costly exploratory detours efficiently.

Furthermore, the inherent heterogeneity of chiplet systems creates a significant optimization challenge. 
Integrating various IPs from different process nodes establishes highly complex and non-linear relationships between architectural parameters and performance metrics.
Traditional Design Space Exploration (DSE) methods, while mathematically robust, struggle to handle this surging complexity efficiently.
On one hand, chiplet-specific frameworks like RapidChiplet\cite{rapidchiplet} and HISIM\cite{hisim} rely on heuristic, formula-based analytical models, which do not provide DSE algorithmic optimization for automated architecture search.
On the other hand, advanced search strategies like XGBoost $\cite{Xgboost}$ and MCT-Explorer $\cite{Mcts}$ primarily focus on optimizing the exploration strategy itself, treating the system as a ``black box'' and relying on statistical correlations rather than underlying architectural logic.
In the context of chiplet design, where cycle-accurate simulations are computationally expensive, these data-driven methods face a severe small-sample constraint.
So, incorporating microarchitectural expertise is important for improving sample efficiency, especially when costly cycle-accurate simulations provide only limited data.
However, existing approaches often introduce such domain knowledge into DSE in a relatively rigid manner.
For example, BOOM-Explorer~\cite{boom} biases initial sampling toward parameters considered important by experts, while SoC-Tuner~\cite{soc-tuner} uses parameter importance mappings to prune or transform the design space.

Recent advances in LLMs offer a new solution to this problem.
By leveraging domain knowledge, LLMs can interpret complex architectural parameters of heterogeneous chiplets and guide optimization in a knowledge-driven manner \cite{llm, cot}.
Recent works have also started to apply LLMs to DSE.
For instance, LLAMBO \cite{LLAMBO} and LEMOE \cite{lemoe} integrate LLMs into Bayesian optimization to accelerate the process via intelligent warm-starting, surrogate modeling, and candidate sampling.
LLM-DSE \cite{LLM-DSE} proposes a multi-agent framework for HLS design, ChatDSE \cite{ChatDSE} uses a zero-shot conversational approach to explore microarchitectures, and Yao \etal \cite{high} use an LLM as both a feature extractor and an autonomous agent for HLS directive optimization.
Despite their promise, these methods rely on a flawed assumption: the LLM is treated as an oracle that can predict the performance of any design and recommend the optimum.
In practice, LLMs are unreliable for precise numerical optimization and may generate biased recommendations when faced with unfamiliar chiplet designs, limiting their effectiveness and generalization.


To address these challenges, we propose a framework for the architectural development and optimization of heterogeneous chiplets.
We first extend the 2.5D simulator LegoSim \cite{LegoSim} to support 3D architectures by introducing a graph-based topology representation and routing.
We then integrate the domain knowledge and reasoning of an LLM with the strategic exploration of Bayesian optimization, thereby improving the efficiency of DSE for heterogeneous chiplets.
Our contributions are summarized as follows:
\begin{itemize}[leftmargin=2em]
    \item We propose an LLM-augmented Bayesian optimization framework for the optimization of heterogeneous 2.5D/3D chiplets, where LLM-generated adaptive directional policies complement BO-based numerical search.
    \item We introduce an adaptive policy reasoning mechanism and exploration gain-aware fine-tuning method to improve the quality of LLM-generated search policies.
    \item We propose a policy-guided Bayesian optimization using a novel P-EHVI acquisition function, which leverages LLM policies to focus the exploration on more promising regions.
    \item We conduct experiments on both 3D and 2.5D chiplet datasets. Results demonstrate that our framework increases hypervolume by $13.3\%$ compared to the previous method on the primary 3D dataset, and achieves a $9.3\%$ improvement in hypervolume on the 2.5D dataset without retraining, demonstrating transferability across related CPU-GPU chiplet tasks.
\end{itemize}
\section{Preliminaries}
\label{sec:prelim}

\subsection{Chiplet-based Heterogeneous Systems}
Chiplet technology is an advanced paradigm for heterogeneous integration.
This approach decomposes a conventional monolithic SoC into multiple independent chips, which enables flexible integration across technology nodes and IP from different vendors and yields substantial benefits in performance, cost, and development time.
Our work focuses on a representative heterogeneous 3D/2.5D chiplet system composed of a host CPU and multiple GPUs interconnected by an interposer for high-efficiency on-package communication.
To perform architectural-level simulation, we use the LegoSim simulator \cite{LegoSim}, which integrates multiple standalone simulators, including Gem5\cite{gem5} for CPUs and GPGPU-Sim\cite{gpgpu-sim} for GPUs, via a dedicated communication protocol.

\subsection{Adaptive Policy for DSE}

\begin{figure*}[htb]
    \centering
    \includegraphics[width=0.9\linewidth]{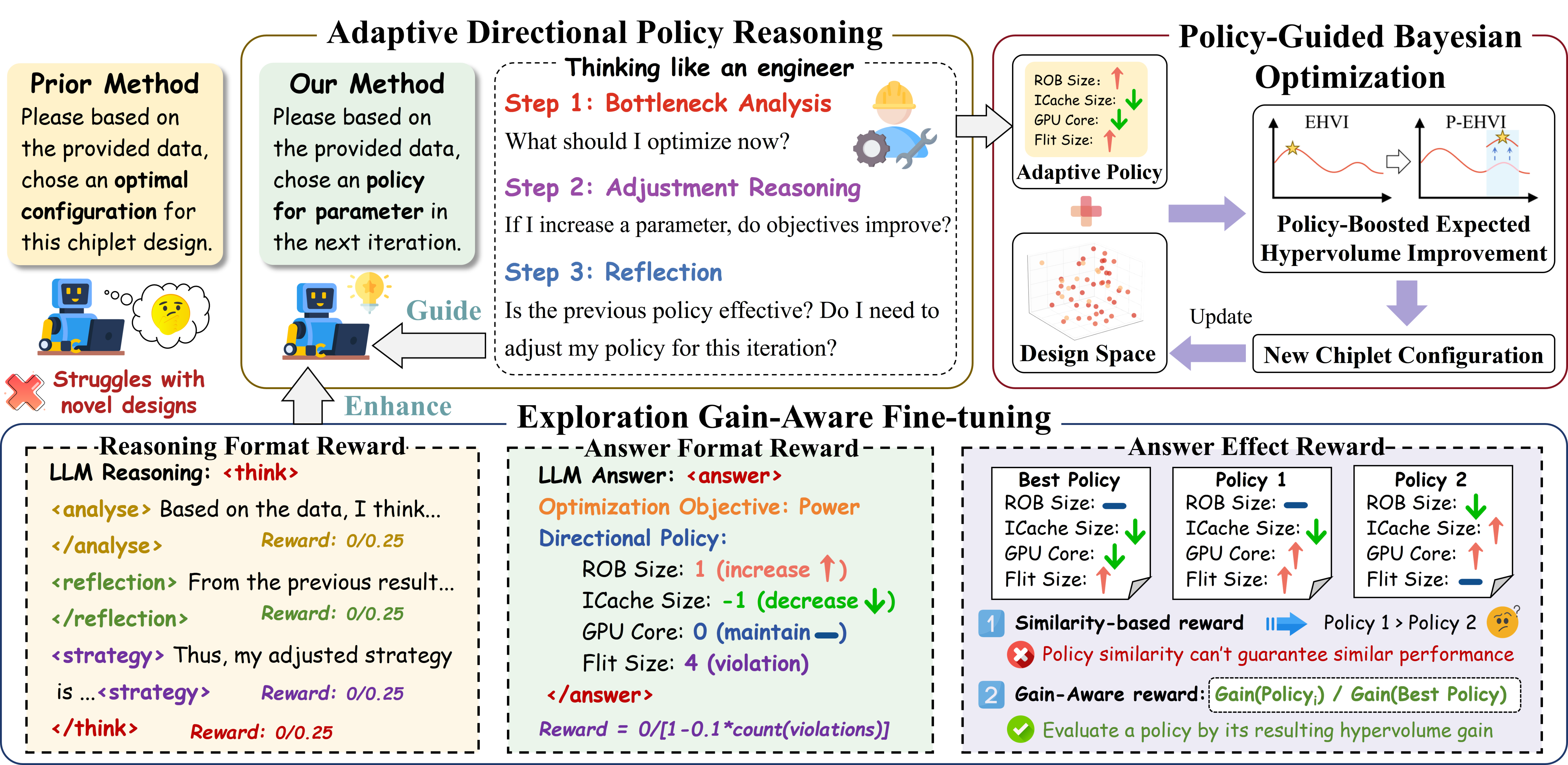}
    \Description{Workflow of the causal policy-driven LLM–BO synergistic framework. Our causal policy synthesis chain-of-thought and exploration gain-aware fine-tuning method train the LLM to generate high-quality causal policies that guide BO.}
    \caption{Workflow of the LLM-augmented BO framework. Adaptive directional policy reasoning and exploration gain-aware fine-tuning enable the LLM to generate effective directional policies that guide BO.}
    \label{fig:worckflow}
\end{figure*}

Traditional DSE methods, such as Bayesian optimization~\cite{bo_ori} and genetic algorithms~\cite{genetic}, typically treat the design evaluator as a black box and infer promising configurations from observed samples.
Under limited simulation budgets, however, sparse observations may be insufficient to efficiently identify beneficial parameter adjustments. In contrast, microarchitecture engineers use domain knowledge to identify performance bottlenecks and reason about the likely effects of adjusting related design parameters.

Inspired by this process, we define an adaptive directional policy $\pi:\mathcal{S}\rightarrow\mathcal{A}$, which maps the current exploration state $\bm{s}\in\mathcal{S}$ to a directional action $\bm{a}\in\mathcal{A}$.
The state summarizes the current Pareto set, the selected anchor design, and feedback from previous iterations, while each element of $\bm{a}$ specifies whether the corresponding parameter should be increased, decreased, or maintained.
Unlike fixed heuristic rules, the LLM uses Adaptive Directional Policy Reasoning (ADPR) to regenerate the policy at each iteration based on the current exploration state and previous feedback. The resulting policy adapts to changing bottlenecks and provides a qualitative search prior that biases BO toward promising parameter adjustments without replacing its uncertainty-based candidate selection.



\subsection{Problem Formulation}
The primary goal of Chiplet DSE is to identify the Pareto front, a set of non-dominated solutions that represents the optimal trade-offs among competing objectives.
The quality of a discovered Pareto front $\Omega$ is quantified by the following metrics:
\begin{mydefinition}[Hypervolume]
\textnormal{
    Hypervolume (HV) captures both the convergence and diversity of the solution set by measuring the volume of the objective space it dominates\cite{HyperVolume}, bounded by a worst-case reference point $f^{ref}$:
    \begin{equation}
    \text{HV}(f^{ref}, \Omega) = \text{Volume}\left(\bigcup_{\bm{x} \in \Omega} \prod_{i=1}^{m} [f_i(\bm{x}), f_i^{ref}]\right),
    \label{eq:hypervolume}
    \end{equation}
    where $\prod$ denotes the Cartesian product of the objective intervals, which defines an $m$-dimensional hyperrectangle.
    }
\end{mydefinition}

\begin{mydefinition}[ADRS]
\textnormal{
    Average distance to reference set\cite{ADRS} evaluates the proximity of found front $\Omega$ to the true Pareto set $\Gamma$:
    \begin{equation}
    \text{ADRS}(\Gamma, \Omega) = \frac{1}{|\Gamma|} \sum_{\gamma \in \Gamma} \min_{\omega \in \Omega} ||\gamma - \omega||_2,
    \label{eq:ADRS}
    \end{equation}
    where $||\cdot||_2$ denotes the Euclidean distance.
    }
\end{mydefinition}

\begin{myproblem}[Multi-Objective heterogeneous Chiplet DSE]
\textnormal{
    The primary objective of chiplet DSE is to identify a Pareto-optimal set of configurations.
    In this work, we optimize power, cycles per instruction (CPI), and interconnect delay while imposing an architectural-level thermal feasibility constraint.
    Given the per-chiplet power estimates and a fixed package floorplan, a lightweight steady-state thermal evaluator estimates the peak temperature $T_{\mathrm{peak}}$ \cite{mfit}.
    Only thermally feasible designs are included in the feasible Pareto set.
    The thermal model is used for early-stage architectural screening rather than signoff-level thermal verification.
}
\end{myproblem}

\section{LLM-Augmented Bayesian Optimization Framework}
\label{sec:flow}


This section presents our LLM-augmented Bayesian optimization framework, as illustrated in \Cref{fig:worckflow}. 
The framework combines Adaptive Directional Policy Reasoning (ADPR), exploration gain-aware fine-tuning, and policy-guided Bayesian optimization.
ADPR enables the LLM to use the current exploration state and simulation feedback to generate adaptive directional policies. The fine-tuning objective improves policy quality, while P-EHVI incorporates these policies as soft search priors into BO.

\subsection{Why LLM-BO Synergy?}
Chiplet DSE aims to identify Pareto designs under a limited simulation budget.
Traditional methods rely primarily on sampled data and usually incorporate prior knowledge through static rules or biased initialization, limiting their effectiveness in sparse-data regimes.
LLMs can interpret design context and generate flexible qualitative search strategies, but they remain unreliable for precise numerical prediction and may introduce systematic search bias.
We therefore combine their complementary strengths: the LLM generates adaptive directional policies that bias the acquisition function toward promising regions, while BO remains the numerical optimization engine and balances exploitation with uncertainty-aware exploration.
This division of roles improves search focus without replacing BO's principled candidate selection.

\subsection{Overall Optimization Flow}

We extend LegoSim from 2.5D to 3D chiplet architectures by introducing graph-based topology modeling and routing, together with a TSV power model.
Based on the enhanced simulator, we construct representative 2.5D and 3D CPU-GPU chiplet design spaces and optimize three objectives: power, cycles per instruction (CPI), and interconnect delay.
After each architectural simulation, the per-chiplet power estimates are mapped to the corresponding package floorplan, and HotSpot \cite{hotspot} estimates the peak steady-state temperature.
Designs exceeding the temperature limit are marked as thermally infeasible and excluded from the feasible Pareto set.

The optimization engine is decoupled from the architectural evaluator, allowing the framework to be extended to other target systems, such as DRAM, NPU, and CIM-based architectures, by integrating the corresponding simulators.

\begin{figure}
    \centering
    \includegraphics[width=0.9\linewidth]{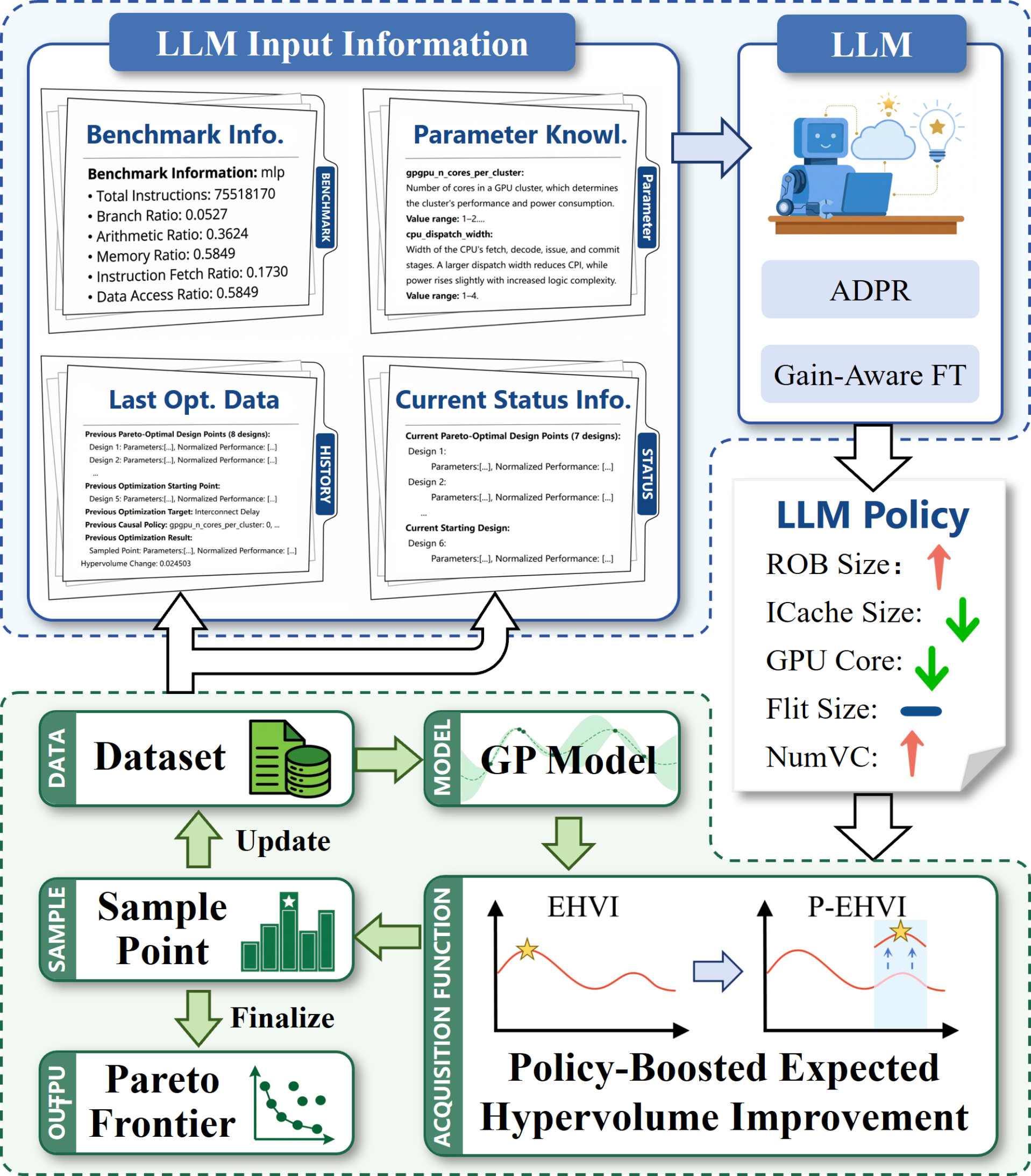}
    \caption{Detailed optimization flow of our LLM-augmented Bayesian optimization framework.}
    \label{fig:opt_flow}
    \Description{Detailed optimization flow of our causal policy-driven LLM-BO synergistic framework.}
\end{figure}

\Cref{fig:opt_flow} shows the detailed workflow of a single optimization iteration within our synergistic DSE framework.
The optimization loop proceeds as follows: first, our exploration gain-aware fine-tuned LLM generates adaptive policies for parameter optimization through engineer-like reasoning.
Second, these policies guide Bayesian optimization to focus on promising subspaces, which enhances DSE efficiency and accelerates convergence.

\subsection{Adaptive Directional Policy Reasoning}

Directly using an LLM to predict optimal configurations is unreliable because chiplet DSE involves complex numerical trade-offs.
In practical design exploration, engineers instead follow an iterative process that identifies performance bottlenecks, reasons about suitable parameter adjustments, and revises the search direction according to observed outcomes.

Inspired by this process, we propose Adaptive Directional Policy Reasoning (ADPR), a structured reasoning procedure that enables the LLM to generate qualitative search guidance rather than directly predict design performance.
At each iteration, ADPR uses benchmark information, the current exploration state, and feedback from the previous iteration to produce an adaptive directional policy. The procedure consists of three stages:\\
\textbf{1) Bottleneck Analysis:}
The LLM analyzes the benchmark information and the current Pareto set $\Omega_t$ to identify the primary bottleneck of the optimization anchor $\bm{x}_t^*$. The anchor is selected as the design with the largest hypervolume contribution:
\begin{equation}
    \bm{x}_t^* =\underset{\bm{x} \in \Omega_t}{\arg\max}\left[\mathrm{HV}(\Omega_t)-\mathrm{HV}(\Omega_t \setminus \{\bm{x}\})\right].
\end{equation}
\textbf{2) Adjustment Effect Prediction:}
Based on the identified bottleneck, the LLM uses parameter descriptions, domain knowledge, and current observations to predict the effects of increasing, decreasing, or maintaining related parameters and determine promising adjustment directions.\\
\textbf{3) Feedback Reflection:}
The LLM revises its reasoning using the previous state $\bm{s}_{t-1}$, directional policy $\bm{a}_{t-1}$, newly evaluated designs, and the resulting hypervolume change. The first ADPR iteration uses feedback from an initial standard BO step.

Finally, ADPR generates an adaptive directional policy $\bm{a}_t$, with each element specifying whether a parameter should be increased, decreased, or maintained.
This policy is updated at each iteration and serves as a qualitative prior for the subsequent BO search.

\subsection{Exploration Gain-Aware Fine-tuning}
We propose an exploration gain-aware fine-tuning method to align the LLM's reasoning with the adaptive directional policy reasoning procedure and improve directional policy generation.
Conventional supervised fine-tuning (SFT) often encourages rigid imitation of training examples~\cite{sft}, limiting the model's ability to discover effective reasoning strategies for complex DSE tasks.
We therefore adopt group relative policy optimization (GRPO)~\cite{GRPO} with an exploration gain-aware reward that aligns model training with the objective of improving the final Pareto front.
The resulting reward, $R_{\text{total}}$, evaluates the LLM output in terms of structural validity and exploration effectiveness:
\begin{equation}
    R_{\text{total}} = \alpha R_{\text{think}} + \beta R_{\text{answer}} + \gamma R_{\text{gain}}.
\end{equation}
We set the weights $\alpha, \beta, \gamma$ to 0.2, 0.2, and 0.6, respectively.
The components of the reward function are detailed below:\\
\textbf{1) Reasoning Format Reward ($R_{\text{think}}$):} This reward encourages compliance with the structured ADPR procedure.
The reasoning is enclosed within \texttt{<think>} tags and organized using \texttt{<analysis>}, \texttt{<reflection>}, and \texttt{<strategy>} tags:
\begin{equation}
    R_{\text{think}} = \frac{|T_{\text{correct}}|}{|T_{\text{req}}|},
\end{equation}
where $T_{\text{req}}$ is the set of all required tags, and $T_{\text{correct}} \subseteq T_{\text{req}}$ is its subset containing tags correctly formatted in the model's response.\\
\textbf{2) Answer Format and Validity Reward ($R_{\text{answer}}$):} This reward ensures that the adaptive directional policy generated by the LLM is correctly formatted and valid for downstream BO.
We first check if the output policy is enclosed by <answer> tags.
Then, we calculate the proportion of parameters that are assigned a valid directional instruction, defined as:
\begin{equation}
R_{\text{answer}} =
\left\{
  \begin{array}{cl}  
    0, & \text{Incorrect Format} \\
    \frac{|P_{\text{valid}}|}{|P|}, & \text{Otherwise}
  \end{array}
\right.
\end{equation}
where ${P}$ is the set of all design parameters and ${P}_{\text{valid}} \subseteq {P}$ is the subset of parameters with a valid directional instruction.\\
\textbf{3) Exploration Gain Reward ($R_{\text{gain}}$):} This reward component is the core of our mechanism, as it directly quantifies a policy's effectiveness by evaluating its expected contribution to the Pareto front.
First, based on the current optimization anchor point $\bm{x}_t^*$, we identify the set of all possible valid policy combinations $\mathcal{A}$.
For each candidate policy $\bm{a} \in \mathcal{A}$, we define its corresponding constrained subspace $\mathbb{D}_{\bm{a}}=\{\mathbb{X}_{\bm{a}},\mathbb{Y}_{\bm{a}}\}$.
For thermally infeasible configurations, we replace their objective values with penalized worst-case values, ensuring that they do not contribute to the theoretical maximum hypervolume improvement (TMHI) within the subspace:
\begin{equation}
    \text{TMHI}(\bm{a}) =  \text{HV}(\Omega_t \cup \mathbb{Y}_{\bm{a}}) - \text{HV}(\Omega_t).
\end{equation}
Finally, the exploration gain score for the LLM-generated policy $\bm{a}_{\text{LLM}}$ is given by its normalized TMHI:
\begin{equation}
    R_{\text{gain}} = \frac{\text{TMHI}(\bm{a}_{\text{LLM}})}{\max_{\bm{a'} \in \mathcal{A}} \text{TMHI}(\bm{a'})}.
\end{equation}

\subsection{Policy-Guided Bayesian Optimization}

To translate the LLM-generated directional policy into candidate selection, we propose a policy-guided Bayesian optimization method.
As illustrated in \Cref{fig:opt_flow}, the evaluated designs are used to train a Gaussian Process (GP) surrogate, while the LLM-generated policy provides directional guidance for the acquisition function.

Given the current anchor point $\bm{x}^{*}$ and directional policy $\bm{a}$, we construct a policy-consistent subspace $\mathcal{X}_{\bm{a}}$.
We then extend Expected Hypervolume Improvement (EHVI)~\cite{EHVI} to Policy-Boosted EHVI (P-EHVI), which softly increases the acquisition values of candidates within $\mathcal{X}_{\bm{a}}$:
\begin{equation}
\label{eq:pehvi}
\begin{aligned}
    \text{EHVI}(\bm{x}|\mathbb{D}) &= \mathbb{E}_{GP}[\text{HV}(\mathbb{D} \cup \{\bm{x}\}) - \text{HV}(\mathbb{D})], \\
    \text{P-EHVI}(\bm{x}|\mathbb{D}) &= \text{EHVI}(\bm{x}|\mathbb{D})  \cdot \left(1 + \lambda \cdot I(\bm{x} \in \mathcal{X}_{\bm{a}}\right)),
\end{aligned}
\end{equation}
where $I(\cdot)$ is the indicator function and $\lambda$ controls the policy boost.
We set $\lambda=8$ according to the empirical scale of EHVI scores.
In our experiments, potentially valuable candidates, including those with high uncertainty or large expected objective improvement, typically have EHVI scores 15 to 20 times higher than clearly low value candidates.
Since $\lambda=8$ corresponds to a ninefold boost, poor candidates within $\mathcal{X}_{\bm{a}}$ generally cannot dominate strong candidates outside it, allowing BO to escape misleading LLM guidance.
When $\mathcal{X}_{\bm{a}}$ contains promising candidates, the uniform boost preserves their relative EHVI ranking while prioritizing the suggested region.
Thus, P-EHVI introduces soft policy guidance without sacrificing BO's exploration capability.

\section{Experiment}
\label{sec:exp}

\subsection{Experimental Setup}

\textbf{Target Architectures and Benchmarks:} As shown in \Cref{fig:flow}, we evaluate two heterogeneous chiplet architectures\cite{heterogeneous_chiplets}, both comprising an ARM-based host CPU and PTX-480 GPUs: a 1-CPU+3-GPU system with 3D packaging running a compute-intensive MLP inference benchmark, and a 1-CPU+2-GPU system using 2.5D packaging, tasked with a parallel matrix multiplication benchmark.
\begin{table}[t]
    \renewcommand{\arraystretch}{1}
    \centering
    \caption{Configurable parameters of the heterogeneous CPU-GPU chiplet architectures.}
    \label{tab:params}
    \resizebox{0.9\linewidth}{!}{
    \begin{tabular}{c|l|c}
    \hline
    Module    & \multicolumn{1}{c|}{Parameters}  & Candidates       \\ \hline
    \multirow{2}{*}{\texttt{GPU}} & CoresPerCluster & 1, 2 \\
    & CacheModel & 0, 1 \\ \hline
    \multirow{6}{*}{\texttt{CPU}} & DispatchWidth & 1, 2, 3, 4 \\
    & ROBSize & 32, 64, 96, 128 \\
    & ICacheSize & 32, 64 \\
    & ICacheAssociativity & 4, 8 \\
    & DCacheSize & 32, 64, 128 \\
    & DCacheAssociativity & 4, 8 \\ \hline
    \multirow{3}{*}{\texttt{Network}} & RoutingBuffer & 6, 12 \\
    & FlitSize & 2, 4, 8 \\
    & NumVC & 2, 4 \\
    \hline
    \end{tabular}
    }
\end{table}


\textbf{Design Space and Simulation Environment:} As shown in \Cref{tab:params}, our experimental setup defines a heterogeneous chiplet design space with 11 parameters (6 for CPU, 2 for GPU, and 3 for interconnect), totaling 18,432 configurations.
Each configuration is evaluated for power, CPU CPI, and interconnect delay using our enhanced 2.5D/3D chiplet simulation platform.
From the suite of simulators integrated within the platform, we select Sniper \cite{sniper} for CPU simulation and GPGPU-Sim \cite{gpgpu-sim} for GPU simulation.
We use HotSpot \cite{hotspot} as a lightweight steady-state thermal evaluator.
Per-chiplet power estimates are mapped to fixed floorplan and package-stack models for steady-state thermal evaluation.
Each individual design evaluation necessitates approximately 10 minutes of simulation time; consequently, the generation of the entire chiplet dataset consumed a total of 6,144 CPU core-hours.


\textbf{Datasets:} We randomly sample 2,000 configurations from the 3D chiplet design space and construct 6,000 prompts from exploration histories, each consisting of 10 to 40 evaluated designs.
To cover diverse optimization outcomes, the histories follow a $30\%/40\%/30\%$ distribution of high, moderate, and zero hypervolume gains.
Each scenario is paired with a reference directional policy derived from the offline dataset.
After deduplication and filtering, the prompts are divided into 4,800 training and 1,200 validation samples.

\textbf{LLM Fine-Tuning:} We fine-tune Qwen2.5-7B-Instruct~\cite{qwen2} using GRPO with the proposed exploration gain-aware reward function. This one-time offline training is conducted on eight NVIDIA A800 GPUs and takes approximately 20 hours. It aligns the model with the structured policy reasoning procedure and improves its ability to generate valid directional policies for guiding subsequent BO-based exploration.

\begin{table*}[tb!]
    \scriptsize
    \renewcommand{\arraystretch}{1}
    \centering
    \setlength{\tabcolsep}{1mm} 
    \caption{Comparison of DSE performance against state-of-the-art methods. The ``Qwen'' and ``GPT'' headers refer to the Qwen-235B and GPT-5.4 models.}
    \label{tab:result}
    \resizebox{1\textwidth}{!} {
    \begin{tabular}{cccccccccc}
        \toprule
        \textbf{Dataset} & \textbf{Metric} & \textbf{XGBoost} & \textbf{\makecell{BOOM\\Explorer}} & \textbf{\makecell{MCT\\Explorer}} & \textbf{\makecell{LLAMBO\\(w/Qwen)}} & \textbf{\makecell{LLAMBO\\(w/GPT)}} & \textbf{\makecell{LLM-DSE\\(w/Qwen)}} & \textbf{\makecell{LLM-DSE\\(w/GPT)}} & \textbf{Ours} \\
        \midrule
             
 \multirow{2}{*}{3D chiplet} & HV $\uparrow$ & 0.7509 & 0.7466 & 0.7376 & 0.7881 & 0.7549 & 0.8210 & 0.8209 & \textbf{0.8356 ({\color{red}\textuparrow 13.3\%})} \\  
                          & ADRS $\downarrow$ & 0.1116 & 0.1028 & 0.1378 & 0.1129 & 0.0986 & 0.1405 & 0.1417 & \textbf{0.0983 ({\color{red}\textdownarrow 28.7\%})} \\ \midrule
 \multirow{2}{*}{2.5D chiplet} & HV $\uparrow$ & 0.7502 & 0.7630 & 0.7366 & 0.7732 & 0.7809 & 0.7564 & 0.7638 & \textbf{0.8051 ({\color{red}\textuparrow \ 9.3\%})} \\ 
                          & ADRS $\downarrow$ & 0.1239 & 0.0859 & 0.1239 & 0.1175 & 0.1095 & 0.1315 & 0.1289 & \textbf{0.0835 ({\color{red}\textdownarrow 32.6\%})} \\ 
        \bottomrule
\end{tabular}
}
\end{table*}

\subsection{Comparison with SOTA Methods}

\begin{figure}
    \centering
    \includegraphics[width=1\linewidth]{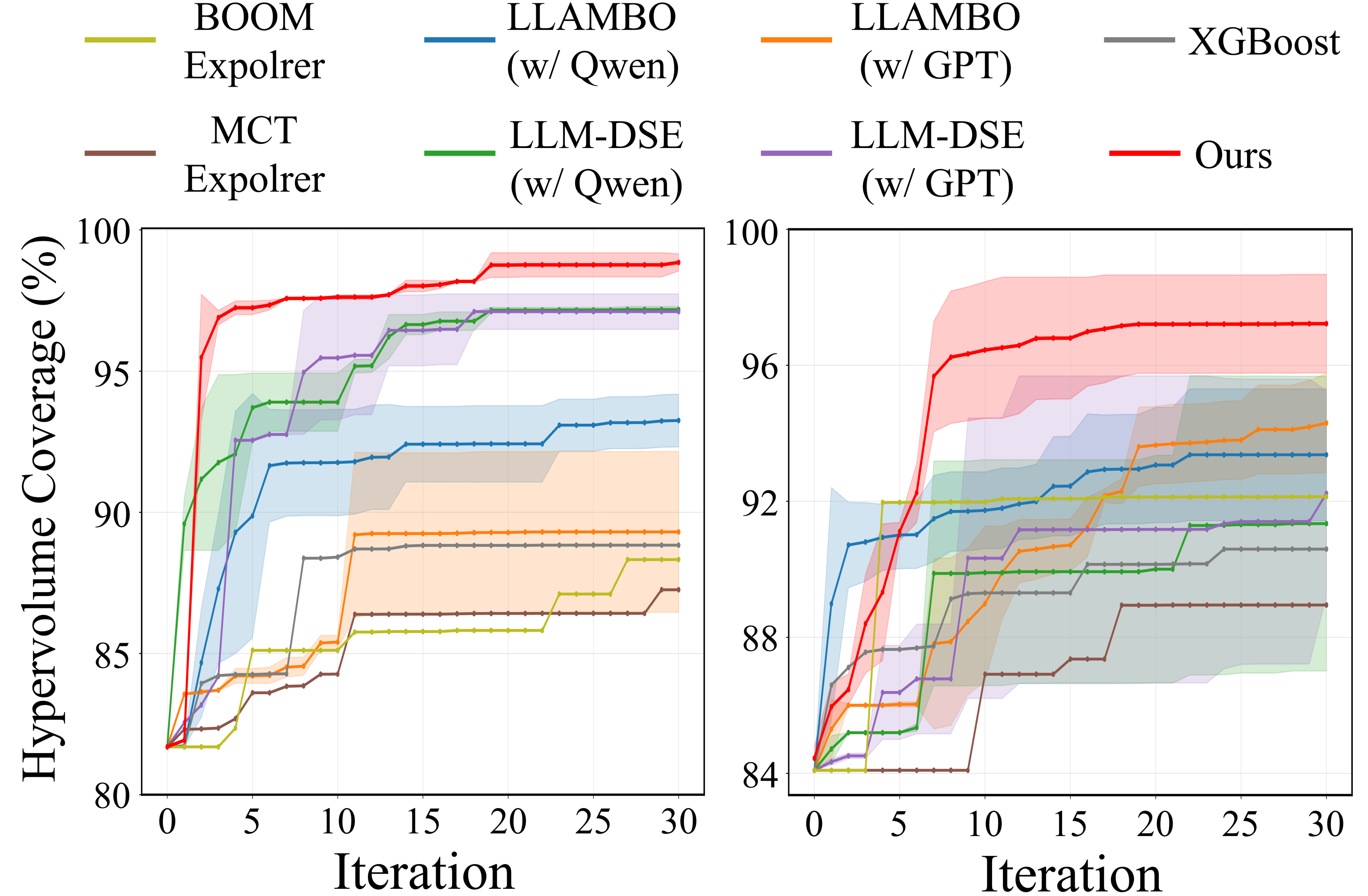}
    \Description{Comparison of hypervolume coverage for different methods over time on two datasets, left is 3D chiplet with MLP, and right is 2.5D chiplet with matrix multiplication.}
    \caption{Comparison of hypervolume coverage for different methods over time on two datasets, left is 3D chiplet with MLP, and right is 2.5D chiplet with matrix multiplication.}
    \label{fig:hv}
\end{figure}

We compare our method with three black-box DSE baselines, XGBoost~\cite{Xgboost}, BOOM-Explorer~\cite{boom}, and MCT-Explorer~\cite{Mcts}, and two LLM-based methods, LLAMBO~\cite{LLAMBO} and LLM-DSE~\cite{LLM-DSE}.
The LLM-based baselines use strong API-based models, including GPT-5.4 and Qwen-235B.
All methods are evaluated using five random seeds, with the same 10 initial designs and 30-iteration budget shared across methods for each seed.
To account for LLM output stochasticity, each LLM-based method is additionally repeated five times per seed, and the resulting HV and ADRS values are averaged.


\Cref{tab:result} shows that our method achieves the best average HV and ADRS on both chiplet design spaces.
The 3D task evaluates in-domain policy-guided exploration, whereas the 2.5D task is excluded from fine-tuning and used solely for transfer evaluation.
Its strong performance without retraining demonstrates transferability across related architectures.
We observe that pure LLM-driven search may achieve competitive HV but suffer from poor ADRS due to premature concentration on locally promising regions, whereas conventional black-box methods provide broader but less efficient exploration.
Our framework combines LLM-generated directional guidance with BO's uncertainty-aware search, improving both convergence and Pareto-front coverage.

\begin{figure}
    \centering
    \includegraphics[width=1\linewidth]{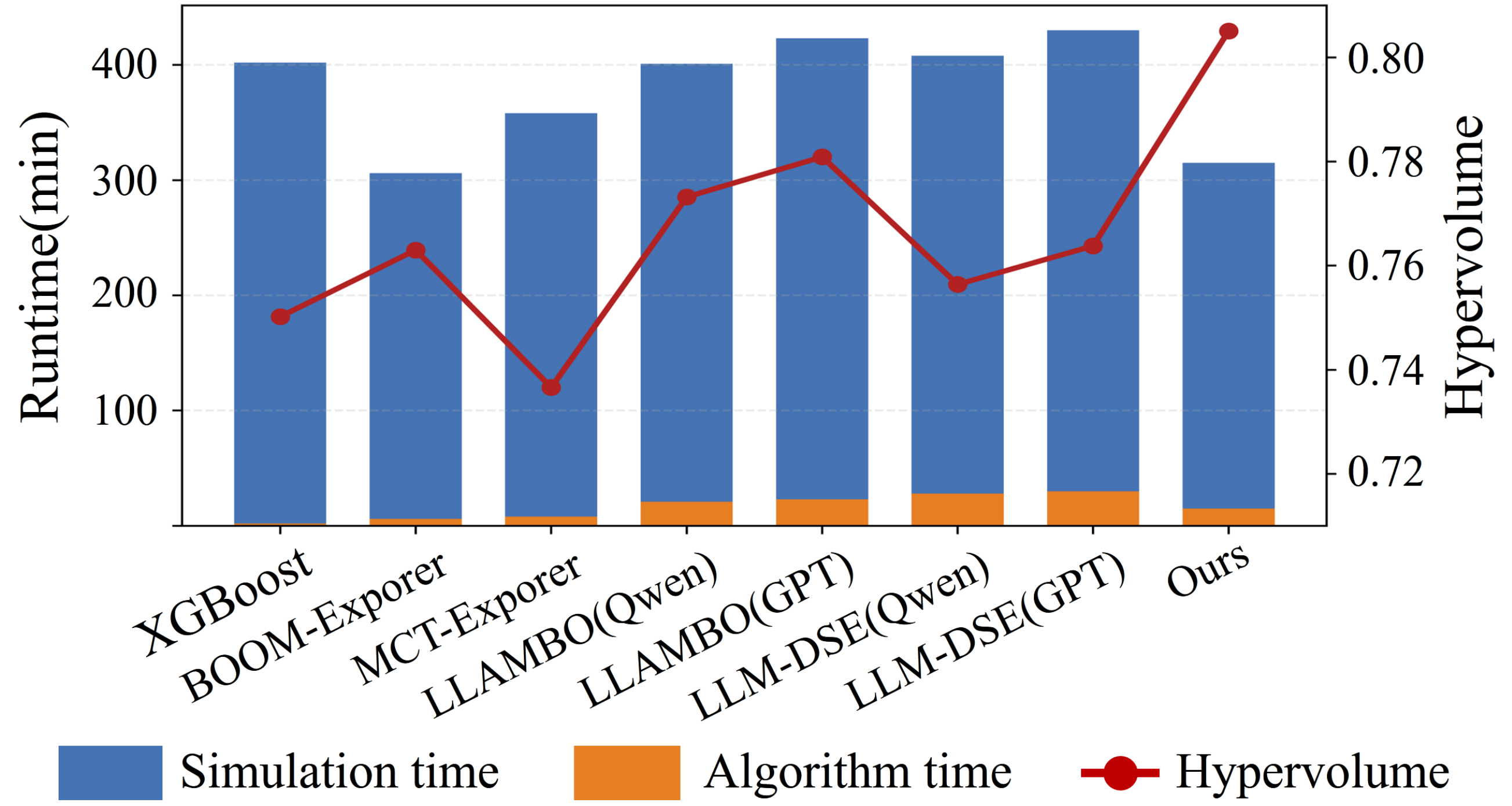}
    \caption{Comparison of runtime distribution and hypervolume across different methods on the MLP benchmark.}
    \Description{Comparison of runtime distribution and hypervolume across different methods on the MLP benchmark.}
    \label{fig:time}
\end{figure}

\Cref{fig:hv} shows that our method achieves 97.22\% of the reference hypervolume derived from the complete 18,432-configuration design space during the early exploration stage on the 3D chiplet dataset, outperforming the baselines.
This result indicates that adaptive directional policies help BO prioritize promising subspaces and avoid unproductive evaluations.
A similar early-stage advantage is observed on the related 2.5D architecture and workload without retraining, demonstrating effective transfer of ADPR-based search guidance across related CPU-GPU chiplet designs.

\Cref{fig:time} compares the online runtime of each method, with simulation and optimization costs reported separately.
Simulation dominates the runtime, making the additional LLM inference overhead relatively small.
The one-time offline fine-tuning cost is reported separately because it is not repeated during exploration and can be amortized across related DSE tasks.
This adaptation enables the compact Qwen2.5-7B model to achieve better task-specific exploration performance than baselines using substantially larger API models.
Overall, our method obtains a higher hypervolume with a shorter online runtime.

\begin{table}[t]
    \centering
    \caption{Thermal feasibility of designs from our method.}
    \label{tab:thermal}
    \resizebox{\columnwidth}{!}{
        \begin{tabular}{cccccc}
        \toprule
        Dataset &$T_{\mathrm{lim}}$($^\circ\mathrm{C}$) &Sampled $T_{\mathrm{peak}}$($^\circ\mathrm{C}$) &Pareto $T_{\mathrm{peak}}^{\max}$($^\circ\mathrm{C}$) \\
        \midrule
        2.5D chiplet &85 &[42.3, \ \ 61.8] &57.9 \\
        3D chiplet &85 &[45.6, \ \ 68.4] &63.7 \\
        \bottomrule
    \end{tabular}}
\end{table}

\Cref{tab:thermal} shows that all evaluated and Pareto-optimal designs remain thermally feasible, with the 3D architecture exhibiting higher peak temperatures due to its stacked structure.

\subsection{Ablation Studies}

\begin{table}[tp]
  \centering
  \caption{Ablation study of our framework's key components.}
  \label{tab:ablation_study_styled}
  \setlength{\tabcolsep}{1pt}
  \resizebox{0.49\textwidth}{!}{
    \begin{tabular}{lcc} 
    \toprule
    \textbf{Method} & \textbf{Hypervolume} $\uparrow$ & \textbf{ADRS} $\downarrow$ \\
    \midrule
    Standard BO             & 0.7428 & 0.1272\\
    Base LLM                & 0.7764 & 0.1525\\
    LLM w/ ADPR           & 0.8117 & 0.1486\\
    LLM w/ Sim-FT           & 0.7979 & 0.1610\\
    LLM w/ Sim-FT + ADPR & 0.8247 & 0.1247\\
    \rowcolor[HTML]{D9EAF7} \textbf{Ours (Gain-Aware FT + ADPR)} &\textbf{0.8356} &\textbf{0.0983} \\
    \bottomrule
  \end{tabular}
  }
\end{table}



We conduct an ablation study on the 3D chiplet dataset to isolate the contributions of adaptive directional policy reasoning and exploration gain-aware fine-tuning.
Our full method is compared against standard BO using EHVI without LLM guidance and four LLM-based variants built on Qwen2.5-7B-Instruct~\cite{qwen2} with P-EHVI.
Among them, similarity-based fine-tuning (Sim-FT) adopts GRPO with a reward that measures the similarity between the generated and reference policies.

As shown in \Cref{tab:ablation_study_styled}, standard BO with EHVI provides a reference for evaluating the overall contribution of LLM-based policy guidance.
The base LLM offers useful but limited directional guidance, whereas ADPR improves policy generation by structuring bottleneck analysis, adjustment effect prediction, and feedback reflection.
Moreover, exploration gain-aware fine-tuning outperforms Sim-FT by rewarding the potential exploration gain of a policy rather than imitation of a single reference policy, enabling the LLM to generate diverse and effective directional policies.
\section{Conclusion}
\label{sec:conclu}

In this paper, we propose an LLM-augmented Bayesian optimization framework for chiplet DSE.
The framework uses adaptive directional policy reasoning to enable the LLM to generate search policies from the current exploration state and simulation feedback, while BO remains responsible for uncertainty-aware candidate selection.
Exploration gain-aware fine-tuning further aligns policy generation with the Pareto optimization objective.
Experiments on 2.5D and 3D chiplet design spaces demonstrate improved exploration efficiency over representative baselines and promising generalization across related architectures.

\clearpage

\bibliographystyle{ACM-Reference-Format}
\bibliography{sample-base}

\end{document}